\documentclass[10pt,twocolumn,letterpaper]{article}

\usepackage{cvpr}              

\usepackage{multirow}
\usepackage{xcolor}
\definecolor{ForestGreen}{RGB}{34, 139, 34}
\definecolor{WildStrawberry}{RGB}{255, 67, 164}
\definecolor{Maroon}{RGB}{128, 0, 0} 
\definecolor{Mahogany}{RGB}{192, 64, 0} 
\usepackage{pifont}
\newcommand{\cmark}{\color{ForestGreen}{\ding{52}}}%
\newcommand{\xmark}{\color{Mahogany}{\ding{56}}}%

\usepackage[accsupp]{axessibility}
\usepackage{appendix}
\definecolor{cvprblue}{rgb}{0.21,0.49,0.74}
\usepackage[pagebackref,breaklinks,colorlinks,allcolors=cvprblue]{hyperref}

\def\paperID{3856} 
\def\confName{CVPR}
\def\confYear{2025}

\title{CAP-Net: A Unified Network for 6D Pose and Size Estimation of Categorical Articulated Parts from a Single RGB-D Image}

\author{Jingshun Huang$^{1*}$\quad Haitao Lin$^{1*}$\quad  Tianyu Wang$^{1}$\quad Yanwei Fu$^{1}$\quad  Xiangyang Xue$^{1}$ \quad  Yi Zhu$^{2}$\\
\\ \vspace{-0.2in}
$^1$Fudan University \qquad $^2$Huawei, Noah’s Ark Lab
}
\begin{document}
\maketitle
{
  \renewcommand{\thefootnote}%
    {\fnsymbol{footnote}}
  \footnotetext[1]{indicates equal contributions.}
  \footnotetext[1]{Prof. Yanwei Fu is also with Institute of Trustworthy Embodied Al, and the School of Data Science, Fudan University}
}

\begin{abstract}
\vspace{-0.15in}

This paper tackles category-level pose estimation of articulated objects in robotic manipulation tasks and introduces a new benchmark dataset.
While recent methods estimate part poses and sizes at the category level, they often rely on geometric cues and complex multi-stage pipelines that first segment parts from the point cloud, followed by Normalized Part Coordinate Space (NPCS) estimation for 6D poses. These approaches overlook dense semantic cues from RGB images, leading to suboptimal accuracy, particularly for objects with small parts.
To address these limitations, we propose a single-stage \textbf{Net}work, CAP-Net, for estimating the 6D poses and sizes of \textbf{C}ategorical \textbf{A}rticulated \textbf{P}arts. This method combines RGB-D features to generate instance segmentation and NPCS representations for each part in an end-to-end manner.
CAP-Net uses a unified network to simultaneously predict point-wise class labels, centroid offsets, and NPCS maps. A clustering algorithm then groups points of the same predicted class based on their estimated centroid distances to isolate each part. Finally, the NPCS region of each part is aligned with the point cloud to recover its final pose and size.
To bridge the sim-to-real domain gap, we introduce the RGBD-Art dataset, the largest RGB-D articulated dataset to date, featuring photorealistic RGB images and depth noise simulated from real sensors. 
Experimental evaluations on the RGBD-Art dataset demonstrate that our method significantly outperforms the state-of-the-art approach. Real-world deployments of our model in robotic tasks underscore its robustness and exceptional sim-to-real transfer capabilities, confirming its substantial practical utility. Our dataset, code and pre-trained models are available on the project page~\footnote{Webpage. \url{https://shanehuanghz.github.io/CAPNet}
}.
\vspace{-0.15in}

\end{abstract}

\section{Introduction}
\label{sec:intro}
\begin{figure}
    \centering
    \includegraphics[width=1.0\linewidth]{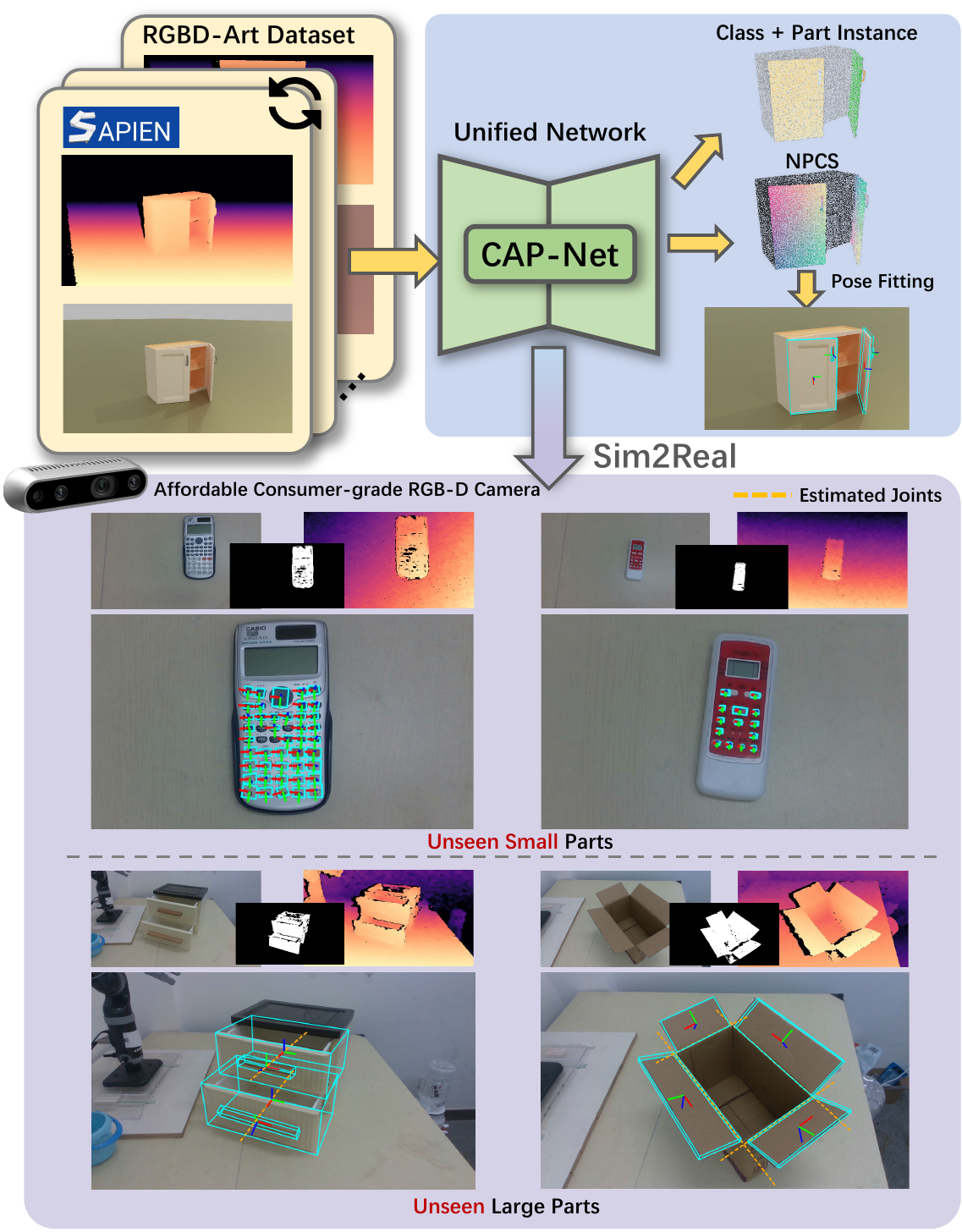}
       \vspace{-0.2in}
    \caption{\textbf{Overview.} CAP-Net is a unified approach for estimating the 6D pose and size of all articulated parts from RGB-D images, requiring only object-level masks instead of part-level ones. The realistic training images in our RGBD-Art dataset allow this synthetic-trained model to effectively adapt to real-world visual perception tasks for robotic manipulation using an affordable RealSense camera.\label{fig:teaser} }
       \vspace{-0.15in}
\end{figure}

Accurately estimating object state is essential for robots to perform diverse grasping and manipulation tasks~\cite{zhang2024category, wen2023foundationpose, cheang2022learning, sun2023language,lin2022know, zhang2024lac}, as shown in Fig.~\ref{fig:teaser}. Recent works~\cite{wang2019normalized,chen2021fs,he2022onepose++} have advanced state estimation for rigid objects, but non-rigid objects like garments~\cite{chi2021garmentnets,xue2023garmenttracking}, fluids~\cite{lin2023pourit,narasimhan2022self}, and articulated parts~\cite{liu2022toward,li2020category} remain challenging due to their complex properties. Articulated parts are particularly difficult because of their rigid, jointed parts, where inaccurate perception can risk damaging delicate joints. Therefore, we focus on improving the perception and estimation of articulated parts to enhance robotic manipulation capabilities.

However, significant challenges remain in accurately perceiving articulated parts, including:
1) \textit{Intra-category part variations.} Novel articulated objects often lack specific 3D CAD models, requiring robust intra-category generalization. For example, estimating the handles of different bucket types necessitates finding shared representations that generalize across various instances within a category.
2) \textit{Cross-category contextual variations.} Articulated parts exhibit diverse contextual configurations across object categories. Unlike category-level rigid object pose methods~\cite{lin2022sar,zhang2024generative,chen2021fs}, which focus on single, consistent shapes, articulated objects involve multiple kinematic parts with unique contexts.
This variability complicates pose and size estimation across different object categories.
3) \textit{Sim-to-real domain gap.} Annotating articulated parts in real-world settings is costly, leading to reliance on synthetic data with more affordable annotations~\cite{geng2023gapartnet, liu2022toward, li2020category, weng2021captra}. However, most synthetic datasets lack photorealistic RGB images and realistic depth data that simulate real sensor capture, creating a sim-to-real gap when transferring pose estimation to real-world applications.

\begin{figure}
    \centering
    \includegraphics[width=1\linewidth]{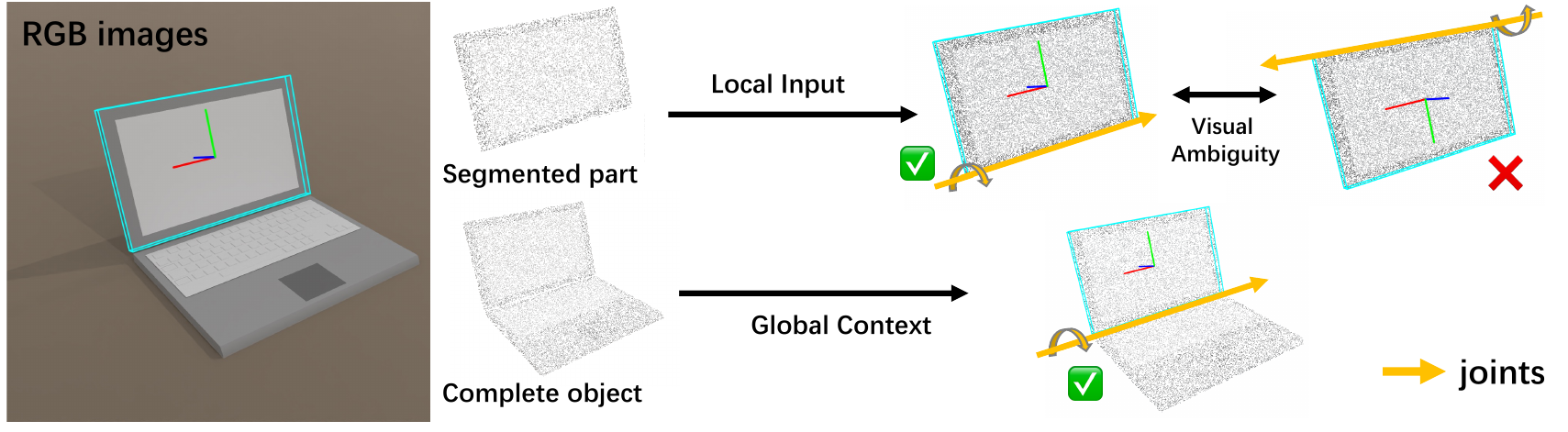}
       \vspace{-0.1in}
    \caption{Comparison of complete-object versus part-based approaches for pose estimation: Recent "first-segment-then-pose" methods, like GAPartNet~\cite{geng2023gapartnet}, rely on processing segmented parts in isolation. This approach can lead to a loss of global context, making it difficult to accurately identify the pose due to visual ambiguity in the input.\label{fig:ambiguity} }
   \vspace{-0.2in}
\end{figure}

Prior point-based methods~\cite{li2020category,geng2023gapartnet,liu2022toward} tackle articulated object pose estimation using Normalized Part Coordinate Space (NPCS) as a standardized frame for category-level mapping from the camera frame. These approaches rely on geometric features but often ignore crucial semantic cues from RGB images, which are vital for precise pose identification. Intra-category methods~\cite{li2020category,liu2022toward} lack cross-category generalization, while GAPartNet~\cite{geng2023gapartnet} extends to cross-category objects but employs a two-stage process—first segmenting parts, then estimating NPCS individually. This cascading approach can accumulate segmentation errors and lose contextual information. For instance, seeing only a laptop’s top part makes it challenging to distinguish the front from the back, while the full object provides cues to resolve this ambiguity (Fig.~\ref{fig:ambiguity}).

To address these challenges, this paper presents a unified, single-shot RGB-D \textbf{Net}work for 6D pose and size estimation of \textbf{C}ategorical \textbf{A}rticulated \textbf{P}arts (CAP-Net), tailored for robotic manipulation tasks. (1) To handle intra-category part variations, we utilize a Normalized Part Coordinate Space (NPCS)~\cite{li2020category, geng2023gapartnet} for each part category. (2) To address cross-category contextual variations, we incorporate category-agnostic features from a pre-trained visual backbones~\cite{fu2024featup, oquab2023dinov2} for part segmentation across unseen object categories. Additionally, our model jointly learns semantic segmentation, centroid offsets, and NPCS mapping. Semantic supervision differentiates part classes, while centroid offsets help distinguish instances within categories, enabling precise localization of novel objects. 
(3) To bridge the sim-to-real gap, we introduce the RGB-D Realistic-Rendering Articulated Object (RGBD-Art) dataset, including photorealistic RGB images and realistic depth images that simulate sensor noise, enabling the adaptation of synthetically trained models to real-world scenarios.

Typically, given RGB-D images, we use the pretrained SAM2~\cite{ravi2024sam} encoder to extract dense features, while FeatUp~\cite{fu2024featup} refines SE(3)-consistent and category-agnostic local semantic features from DINOv2~\cite{oquab2023dinov2, zhang2024tale,zhang2024lapose} to high resolution. These fine-grained semantic features are crucial for part-level detection and pose representation learning. The dense semantic features are fused with the geometric point cloud in a point-wise manner, similar to~\cite{wang2019densefusion}. Using PointNet++~\cite{qi2017pointnet++}, we optimize rigid part segmentation, instance centroid offsets, and dense NPCS coordinate predictions in an end-to-end manner, enhancing overall pose estimation performance. Finally, we align the NPCS regions of each segmented instance with the real point cloud using the Umeyama algorithm~\cite{umeyama1991least} to recover the final pose and size.

In summary, the main contributions of this paper are:
(1) We introduce the first RGB-D method to estimate semantic labels, point-wise instance centers, and NPCS representations simultaneously in an end-to-end, single-shot process using whole-object input. This enables accurate perception of unseen articulated parts from a single RGB-D image in real-world settings.
(2) We present a large-scale, realistic RGB-D dataset with detailed annotations for articulated pose estimation, effectively bridging the domain gap between synthetic and real-world images at both photometric and geometric levels.
(3)
Our method and dataset facilitate sim-to-real transfer to a physical robot, allowing accurate visual perception of articulated components, especially small parts, using only low-cost RGB-D sensors.
This capability supports precise manipulation of target components, highlighting its practical utility in robotic applications.



\section{Related Work}
\noindent\textbf{Part Instance Segmentation.} 
Part instance segmentation in 3D point clouds is challenging due to the irregular and sparse nature of the data. Existing methods, such as PointNet++~\cite{qi2017pointnet++} and SGPN~\cite{wang2018sgpn}, perform segmentation by learning point-wise features using deep learning. More recent methods, like SoftGroup~\cite{vu2022softgroup}, PointGroup~\cite{jiang2020pointgroup}, and AutoPart~\cite{liu2022autogpart}, focus on segmenting instances using geometric cues from the point cloud. GAPartNet~\cite{geng2023gapartnet} also leverages geometric features to learn domain-invariant representations for generalizing across unseen categories. In contrast, our approach incorporates semantic features from RGB images into the point cloud, enhancing segmentation accuracy and improving generalization across different categories and contextual variations.


\noindent\textbf{Category-level Rigid Object Pose Estimation. }
Rigid object pose estimation deals with objects having a fixed shape, requiring a single static pose in 3D space. Some methods~\cite{tian_shape_2020, chen2020learning, wang2019normalized,wang2024polaris,wang2023wall} estimate pose and size from single-view RGB(-D) images, like NOCS~\cite{wang2019normalized}. 
Other depth-based approaches, such as FS-Net~\cite{chen2021fs}, SAR-Net~\cite{lin2022sar}, HS-Pose~\cite{zheng2023hspose}, and GenPose~\cite{zhang2024generative,zhang2025omni6dpose}, estimate pose by purely learning the object's geometry. Unlike these methods, which are limited to rigid objects, our approach focuses on estimating the pose and size of objects with multiple movable parts. Rigid object methods estimate the entire object without needing to consider global context, relying instead on specifically designed metrics~\cite{mo2022es6d, xiang2017posecnn} to address symmetry-related ambiguities. In contrast, our method incorporates global context to resolve visual ambiguity and handle the complexities of pose estimation for objects with movable components.

\noindent\textbf{Category-level Articulated Object Pose Estimation. } 
Articulated object pose estimation~\cite{li2020category, weng2021captra,liu2022akb,liu2022toward,geng2023gapartnet} focuses on objects made of multiple movable parts, requiring the estimation of both the overall pose and the configuration of individual components. To improve accuracy and generalization, the Articulation-aware Normalized Coordinate Space Hierarchy (ANCSH)~\cite{li2020category} represents articulated objects within a category. 
While notable works like~\cite{jiang2022opd,zeng2021visual} address 5-DoF motion axis estimation, they fall short of providing full 6-DoF pose and 3D size estimation, which are crucial for precise robotic manipulation. Additionally,~\cite{zeng2021visual} relies on per-part segmentation masks, which are more difficult to obtain than whole-object masks. In contrast, our approach uses readily available object-level masks to estimate full 6-DoF pose and 3D size.
The AKB-48~\cite{liu2022akb} dataset provides real-world 3D articulated models with a pipeline for part reconstruction and pose estimation. GAPartNet~\cite{geng2023gapartnet} uses a two-stage approach for domain-generalizable 3D part segmentation and pose estimation, but it suffers from error accumulation and relies on high-quality point clouds from industrial cameras. In contrast, our approach improves pose and size estimation accuracy by incorporating semantic cues from RGB images, which is robust to the noise depth from consumer-grade camera.


\noindent\textbf{Articulated Object Dataset.}Large-scale 3D object datasets like ShapeNet~\cite{chang2015shapenet}, Objaverse~\cite{deitke2023objaverse,deitke2024objaverse}, and OmniObject3D~\cite{wu2023omniobject3d} treat objects as rigid bodies without part definitions, which is inadequate for fine-grained robotic tasks such as opening bottle caps or pressing buttons. These tasks necessitate part-level perception, prompting the creation of 3D part-wise datasets~\cite{mo2019partnet,wang2019shape2motion} for part-based object recognition and manipulation~\cite{paschalidou2021neural,xu2022partafford,yang2021unsupervised}. Further research, such as GAPartNet~\cite{geng2023gapartnet}, uses public repositories like PartNet-Mobility~\cite{xiang2020sapien} and AKB-48~\cite{liu2022akb} to generate point clouds with part annotations. Meanwhile, ReArtMix~\cite{liu2022toward} produces semi-authentic RGB-D images using mixed reality techniques and self-scanning models called ReArt-48. 
These datasets lack photorealistic RGB images and depth maps that mimic real sensor noise, creating a sim-to-real gap. Our RGBD-Art dataset addresses this by offering realistic RGB images and depth maps that include sensor noise.

\section{RGBD-Art Dataset}
\subsection{Definition of Poses and Joints}
Each part category is canonically oriented and normalized to the Normalized Part Coordinate Space (NPCS)~\cite{geng2023gapartnet, wang2019normalized, li2020category}, ensuring a consistent definition of pose and joint parameters. This definition simplifies the network's task, as it only needs to estimate the pose without separately estimating joint parameters. 
In real-world robotics, once the object category and 6D pose are established, predefined joint parameters can be easily queried from the estimated oriented bounding box for manipulation, as shown in~\cite{geng2023gapartnet}.

\subsection{Realistic RGB-D Rendering}
Recent datasets like CArt~\cite{li2020category}, GAPartNet~\cite{geng2023gapartnet}, and ReArtMix~\cite{liu2022toward} have driven progress in articulated object pose estimation but still have key limitations. (1) Their synthetic depth images are idealized and lack the realistic noise in real-world sensors like RealSense or Kinect, creating a sim-to-real domain gap when training solely on synthetic data. GAPartNet attempts to address this by using an industrial RGB-D camera for high-quality depth images in real-world applications, but these cameras are costly and have low frame rates, limiting their use on household robotic platforms. 
(2) Additionally, the non-photorealistic RGB images in these datasets create a significant domain gap due to pixel-level distribution differences, causing most works~\cite{li2020category,geng2023gapartnet, liu2022toward} to prioritize geometric cues from point clouds while neglecting valuable RGB features.

\begin{figure}
    \centering
    \includegraphics[width=0.95\linewidth]{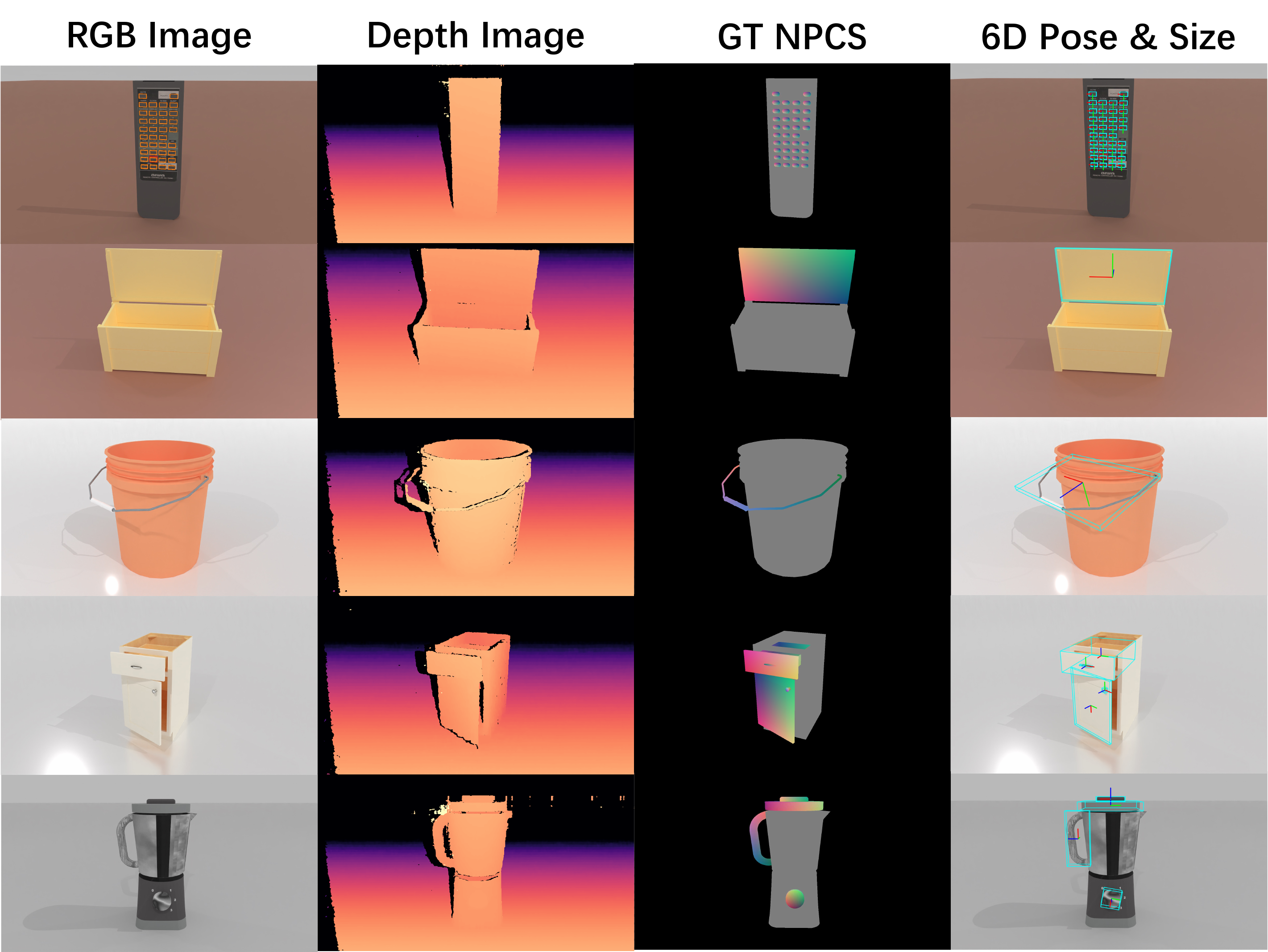}
       \vspace{-0.1in}
    \caption{Exemplars of our RGBD-Art dataset. We show the photo-realistic RGB image, realistic depth images and corresponding ground-truth annotations of NPCS map, 6D pose and size.\label{fig:examplar} }
   \vspace{-0.1in}
\end{figure}

To address limitations in existing datasets, we adopt the techniques in~\cite{xiang2020sapien} to synthesize depth images with realistic sensor noise patterns, simulating an active stereo depth camera similar to the RealSense D415. For RGB images, we employ ray-tracing to achieve photorealism, incorporating domain randomization to vary lighting, textures, and object materials. With these realistic RGB-D images, we can readily obtain accurate annotations for part segmentation and pose. Figure~\ref{fig:examplar} shows the the exemplars of our rendering RGB-D images and the generated ground-truth labels. We also compare our RGBD-Art with the existing synthetic dataset as in Tab.~\ref{tab:dataset}.

\begin{table}[ht]
\footnotesize
\caption{Comparison with other synthetic articulated object datasets. `P-RGB' represents photorealistic RGB images, `R-D' denotes more realistic depth of objects, and `BG' indicates datasets that use real images solely for the background.}
\centering
\begin{tabular}{c|c|c|c|c|c}
\toprule[1pt]
\textbf{Datasets} & \textbf{Obj.} & \textbf{Img.} &  \textbf{Anno.} &\textbf{P-RGB} & \textbf{R-D} \\ \midrule[0.5pt]
ReArtMix ~\cite{liu2022toward} & 48 & 100K & - & \cmark (BG) & \xmark \\
GAPartNet~\cite{geng2023gapartnet} & 1166 & 37K & 272K & \xmark & \xmark \\
RGBD-Art(Ours) & 1045 & 63K & 408K & \cmark & \cmark \\ \bottomrule[1pt]
\end{tabular}

\label{tab:dataset}

\end{table}

\noindent\textbf{Data statistics.}
Building on the objects used in the GAPartNet dataset~\cite{geng2023gapartnet}, we incorporate existing resources from PartNet-Mobility~\cite{xiang2020sapien} that provide URDF models of articulated objects with unified annotations for each part type. 
AKB-48~\cite{liu2022akb} is not used due to its lack of 3D object shapes, which could impact the quality of pose annotations and rendered images.
Our dataset contains 9 categories of articulated part types, including: \textit{line fixed handle}, \textit{round fixed handle}, \textit{hinge handle}, \textit{hinge lid}, \textit{slider lid}, \textit{slider button}, \textit{slider drawer}, \textit{hinge door}, and \textit{hinge knob}. Each instance contains multiple classes of parts. To enhance realism, we place each instance on a flat surface or a desk, simulating typical backgrounds found in household settings.
For each instance, we place it in two random backgrounds and, for each background, randomly sample $60$ camera views to render RGB-D images. In total, we generate 63K images, complete with semantic labels and pose annotations.


\begin{figure*}[t]
    \centering
    \includegraphics[width=1.0\textwidth]{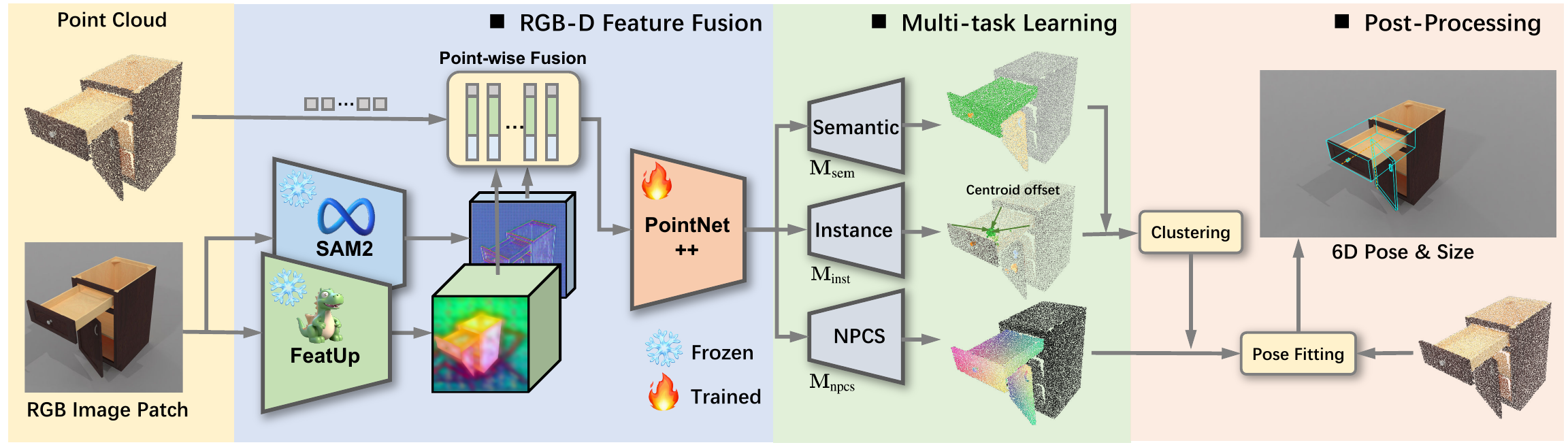}
    \vspace{-0.2in}
    \caption{\textbf{Architecture overview.} 
    CAP-Net uses pretrained vision backbones, SAM2~\cite{ravi2024sam} and FeatUp~\cite{fu2024featup}, to extract dense semantic features, which are then fused with the point cloud in a point-wise manner. The enriched point cloud features are passed into PointNet++~\cite{qi2017pointnet++} for further processing. These features are then used by three parallel modules to predict semantic labels, centroid offsets, and NPCS maps. A clustering algorithm groups points with the same semantic label based on centroid distances to isolate each possible part. Finally, an alignment algorithm matches the predicted NPCS map with the real point cloud to estimate each part’s pose and size.
    \label{fig:pipeline} }
\vspace{-0.15in}
\end{figure*}

\section{Articulated Part Pose Estimation Method}
\noindent \textbf{Task Formulation.} 
Given the RGB-D image patch $(\mathcal{I}, \mathcal{D})$, where $\mathcal{I} \in \mathbb{R}^{H\times W \times 3}$ represents the RGB image and $\mathcal{D} \in \mathbb{R}^{H\times W}$ represents the depth map of the target object, our goal is to estimate the semantic class label $c_i$, the normalized object part coordinate maps $m_i$, and the centroid offsets $o_i$ for each point $i$ in the depth back-projected point cloud $\mathcal{P}$, where $\mathcal{P} \in \mathbb{R}^{N\times 3} $ and $N$ is the number of points in the point cloud.

We start by using the semantic labels to group different part classes. Next, we cluster the centroid offsets $o_i$ to distinguish between different part instances that share the same semantic label. We align the estimated NPCS maps of $k$-th part instance $S_k$ with the corresponding 3D points to compute the 
pose and size parameters $\{\mathbf{R}_k, \mathbf{t}_k, \mathbf{s}_k\}$, where $\mathbf{R}_k \in \text{SO}(3)$, $\mathbf{t}_k \in \mathbb{R}^3$, and $\mathbf{s}_k \in \mathbb{R}^3$. Here, $\text{SO}(3)$ represents the Lie group of 3D rotation.

\noindent \textbf{Architecture Overview.}
As shown in Fig.~\ref{fig:pipeline}, our network first utilizes the pretrained SAM2~\cite{ravi2024sam} and FeatUp~\cite{fu2024featup} encoders to extract semantic features from RGB image patches $\mathcal{I}$, which are then concatenated with the corresponding point cloud $\mathcal{P}$ (Sec.~\ref{sec:extraction}). We then apply the PointNet++~\cite{qi2017pointnet++} backbone to further extract geometric features, which are passed to three modules for multi-task learning. These modules include a semantic part learning module $\mathbf{M}_\text{sem}$ (Sec.~\ref{sec:semantic}), a centroid offset learning module $\mathbf{M}_\text{inst}$ (Sec.~\ref{sec:instance}), and an NPCS learning module $\mathbf{M}_\text{npcs}$ (Sec.~\ref{sec:npcs}), which simultaneously estimate semantic class labels, centroid offsets, and NPCS coordinates for each point.We further cluster points based on their centroid offsets to group those belonging to the same instance, after which part labels are assigned to each instance to filter and locate the NPCS maps for each part. Finally, transformations and scale estimations between the actual points and the NPCS maps produce the 6-DoF pose and 3D size of each part.


\subsection{Feature Extraction and Fusion}
\label{sec:extraction}
Unlike previous point-based methods~\cite{geng2023gapartnet, li2020category, weng2021captra}, our approach integrates RGB images to provide semantic cues for pose estimation. Since our task differs from previous tasks like pose estimation or two-stage articulated pose estimation, we require the network to capture fine-grained features to distinguish multiple small parts within the complete object. Thus, SAM vision encoder provides strong dense feature representations as its excellent open-vocabulary instance segmentation capabilities. Meanwhile, FeatUp extracts SE(3)-consistent and category-agnostic local semantic features similar to DINOv2~\cite{oquab2023dinov2,zhang2024tale,zhang2024lapose}, but preserve higher resolution details.

Particularly, our network begins by utilizing the pre-trained backbone of SAM2~\cite{ravi2024sam} alongside the FeatUp~\cite{fu2024featup} encoder to extract features from the RGB image patch $\mathcal{I}$ separately as illustrated in Fig.~\ref{fig:pipeline}.Specifically, the SAM encoder transforms an image of size $H \times W \times 3$ into a feature representation of size $H \times W \times 96$, while the FeatUp encoder maps the same image into a $H \times W \times 384$ embedding space. Next, we concatenate these feature maps, resulting in an embedding where each pixel corresponds to the 480-dimensional vectors that represent the semantic information of the RGB image at the corresponding location.


Subsequently, we concatenate each RGB feature vector with its corresponding 3D point in a point-wise manner, similar to~\cite{wang2019densefusion}. The 3D points, now enriched with feature vectors, are processed using the pre-trained feature extractors from PointNet++~\cite{qi2017pointnet++} to obtain the densely fused RGB-D features. We then employ three distinct decoders to estimate the corresponding outputs, as illustrated below.

\subsection{Semantic Part Learning}
\label{sec:semantic}
To handle object categories with multiple parts, GAPartNet~\cite{geng2023gapartnet} employs a single network to segment each possible part from the complete point cloud. However, this approach loses the global context needed for accurate pose identification and introduces segmentation errors when each segmented part is processed independently by subsequent networks. To address these limitations, similar to~\cite{li2020category, liu2022toward}, we estimate all semantic labels directly from the input points. Unlike these methods, which assign each part instance a unique class label, even within the same category. They predefine semantic labels for each part, limiting their networks to single-category objects with fixed part counts. In contrast, our approach separately estimates class labels and instance labels (Sec.~\ref{sec:instance}), enabling generalization across categories with variable part structures and counts.


Typically, we introduce a point-wise part semantic segmentation module, $\mathbf{M}_\text{sem}$, into the network to predict semantic labels for each point based on extracted RGB-D features. 
This module integrates dense semantic and geometric cues to streamline the segmentation task, with supervision provided by Focal Loss $FL(\cdot)$, denoted as $\mathcal{L}_\text{sem}$:

\begin{equation}
\mathcal{L}_\text{sem} = \frac{1}{N} \sum_{i=1}^{N} FL(\hat{c}_i, c_i),
\end{equation}

where $\hat{c}_i$ is the predicted label for the $i$-th point, and $c_i$ is the ground-truth class label.


\subsection{Centroid Offset Learning}
\label{sec:instance}

Considering that multiple part instances can share the same semantic label within a complete object, we design the centroid offset Llearning module $\mathbf{M}_\text{inst}$ to predict the centroid of each part instance, thereby enabling the network to differentiation between instances. This module predicts the Euclidean translation offset $\Delta o_i$ of each point to the associated part instance center. The learning process for $\Delta o_i$ is guided by an L1 loss:

\begin{equation}
\label{eqn:Lctr}
    \mathcal{L}_{\text{inst}} = \frac{1}{N} \sum_{i=1}^{N} ||\Delta o_i - \Delta \hat{o}_i || \cdot \mathbb{I}(p_i \in S_k)
\end{equation}
where $N$ represents the total number of seed points on the object's surface, and $\Delta o_i$ is the ground truth translation offset from seed $p_i$ to the instance center. The indicator function $\mathbb{I}$ specifies whether point $p_i$ belongs to the particular instance $S_k$.

Such optimization encourages each point to estimate the potential center of its corresponding part instance, ensuring that estimated centers belonging to the same instance are in close proximity. This facilitates the identification of each instance. Specifically, we further complete the instance segmentation process by first using the semantic labels to filter out background points, and then applying the DBSCAN~\cite{ester1996density} algorithm to cluster the estimated centers of foreground points, thereby distinguishing each instance.

\subsection{NPCS Learning for Pose and Size Estimation}
\label{sec:npcs}


The semantic instance segmentation of the point cloud is handled by the semantic head, $\mathbf{M}_\text{sem}$, and the center offset head, $\mathbf{M}_\text{inst}$. Additionally, we incorporate an NPCS module, $\mathbf{M}_\text{npcs}$, which learns the mapping $\mathcal{P} \rightarrow \mathcal{P}_{\mathbb{C}} \in \mathbb{R}^{N \times 3}$, where $\mathcal{P}$ represents the observed object point cloud, and $\mathcal{P}_{\mathbb{C}}$ denotes the canonical-space point cloud.
The mapping task is formulated as a classification problem by discretizing the NPCS coordinates into 32 bins per axis, which has proven more effective than regression for reducing the solution space~\cite{wang2019normalized}. We supervise NPCS learning with a softmax cross-entropy loss $SCE(\cdot)$, denoted as $\mathcal{L}_\text{npcs}$:
\begin{equation}
\mathcal{L}_\text{npcs} = \frac{1}{3N}\sum_{n=1}^{3}\sum_{i=1}^{N} SCE(\hat{m}_i^n, m_i^n),
\end{equation}
where $m_i^n$ and $\hat{m}_i^n$ are the ground truth and predicted discrete NPCS labels along each axis $n$.

\noindent\textbf{Remark.} 
Some classes exhibit continuous symmetry, so we apply the solution from~\cite{wang2019normalized, geng2023gapartnet} for loss convergence. However, GAPartNet resolves ambiguities in parts like hinged lids or doors by tolerating 180$^\circ$ mirror symmetry due to its reliance on local part inputs. 
In contrast, our method leverages global context, enabling accurate pose direction distinction without relying on symmetry tolerance.

\noindent\textbf{Multi-task Loss.} We supervise the learning of modules $\mathbf{M}_\text{sem}$, $\mathbf{M}_\text{inst}$, and $\mathbf{M}_\text{npcs}$ jointly using a multi-task loss:
$\mathcal{L} = \lambda_1 \mathcal{L}_\text{sem} + \lambda_2 \mathcal{L}_\text{inst} + \lambda_3 \mathcal{L}_\text{npcs}$,
where $\lambda_1$, $\lambda_2$, and $\lambda_3$ are the task-specific weights. Experimental results from the ablation study in Sec.~\ref{sec:ablation} show that jointly training these tasks improves the final performance.

Using the predicted discrete correspondences, we recover the 6D object pose and size through post-processing pose fitting. This is achieved by first applying RANSAC for outlier elimination, followed by the Umeyama algorithm~\cite{umeyama1991least} to estimate the transformation parameters $\{\hat{\rm{s}}, \hat{\mathbf{R}}_k, \hat{\mathbf{t}}_k\}$ that align the predicted canonical-space point cloud  with the estimated segmented part point cloud.


\section{Experiment}

\noindent\textbf{Evaluation Metric.}
Following previous work~\cite{geng2023gapartnet}, we evaluate (1) 3D semantic instance segmentation using AP50, which measures average precision at a 50\% IoU threshold. We also use AP, the average precision across IoU thresholds from 50\% to 95\% in 5\% increments, for overall performance assessment. 
For (2) pose estimation, we assess performance using metrics such as average rotation error \( R_e (^\circ) \), translation error \( T_e (cm) \), scale error \( S_e (cm) \), and translation error along the interaction axis \( d_e (cm) \). Additionally, we measure 3D Intersection over Union (3D mIoU) and calculate accuracy percentages for thresholds of \( 5^\circ \) and \( 5cm \), as well as \( 10^\circ \) and \( 10cm \).

\noindent\textbf{Baselines.} 
We use point-based methods such as PointGroup~\cite{jiang2020pointgroup}, SoftGroup~\cite{vu2022softgroup}, AutoGPart~\cite{liu2022autogpart}, and the state-of-the-art GAPartNet~\cite{geng2023gapartnet} as baseline methods. Note that PointGroup, SoftGroup, and AutoGPart have been adapted for part-based 3D semantic instance segmentation and pose estimation, similar to~\cite{geng2023gapartnet}.

\noindent\textbf{Implementation Details.} We train our model on 4 L20 GPUs for a total of 100 epochs, using a batch size of 4. The initial learning rate is set to 0.001, using a warm-up scheduler for gradual increase at the start of training. Input images are cropped and resized to $640 \times 640$ resolution, and point clouds are randomly sampled to 24,576 points before being processed by the network. The hyperparameters of multi-task loss are empirically set to $\lambda_1=17.5$ , $\lambda_2= 125$ , and $\lambda_3=0.15$ for semantic segmentation, instance offset learning, and NPCS mapping tasks, respectively.

\subsection{Comparison to Baselines}

We compared the accuracy of our method against baseline methods in part segmentation, pose and size estimation, with the results summarized in Tab.~\ref{tab:mainres} and Tab.~\ref{tab:pose_evel}. While our method primarily focuses on pose and size estimation, we also evaluated 3D instance segmentation against baseline methods to highlight the importance of semantic cues from RGB images. The comparison results in Table~\ref{tab:mainres} show that our method outperforms all point-based approaches~\cite{vu2022softgroup,liu2022autogpart,geng2023gapartnet}, achieving an average precision (AP50) of 53.58 in seen object setting and 19.17 in unseen object setting, compared to other state-of-the-art methods. This suggests that incorporating RGB semantic cues is essential for part segmentation, especially for small components like buttons on a remote. Due to noise in depth data, distinguishing each instance using geometric cues can be difficult, but RGB images provide fine details that facilitate the identification of each part.

\begin{table*}[ht]
\footnotesize
\vspace{-0.15in}
\caption{{Results of part segmentation on seen object categories and unseen object categories in terms of per-part-class AP50 (\%), average AP50 (\%)} Ln.=Line. F.=Fixed. Rd.=Round. Hl.=Handle. Ld.=Lid. Bn.=Button. Dw.=Drawer. Dr.=Door. Kb.=Knob.  PG=PointGroup~\cite{jiang2020pointgroup}. SG=SoftGroup~\cite{vu2022softgroup}.
AGP=baseline modified from AutoGPart~\cite{liu2022autogpart}.\label{tab:mainres}}
\vspace{-0.1in}
\centering
\renewcommand\arraystretch{0.9}
\resizebox{.99\textwidth}{!}{
\begin{tabular}
{c|c|ccccccccccc}
\toprule[1pt]
&& Ln.F.Hl. & Rd.F.Hl. & Hg.Hl. & Hg.Ld.& Sd.Ld.  & Sd.Bn  & Sd.Dw. & Hg.Dr. & Hg.Kb. 
& Avg.AP50\\
\midrule[0.5pt]
\multirow{4}{*}{Seen (\%)}  & SG~\cite{vu2022softgroup} &0.02 & 0.07 &0.0 & 10.34 &6.24 &5.56 & 14.32 &3.51 & 0.0 & 4.44 \\ 
                        
 & AGP~\cite{liu2022autogpart} &1.23 & 0.15 &0.01 & 13.24 &10.21 &7.29 &16.45 &7.41 & 0.019 & 6.22 \\ 

& GAPartNet~\cite{geng2023gapartnet} &3.97 &0.26  &0.0 &25.94  &18.41  &12.07   &26.34  &15.15  & 0.038 &11.35 \\ 

& Ours       &\textbf{55.30} &\textbf{16.88} &\textbf{70.71} &\textbf{76.05}&\textbf{93.76} &\textbf{44.06} &\textbf{58.09} &\textbf{51.05} &\textbf{16.23} &  \textbf{53.58} \\ \midrule[0.5pt]
\multirow{4}{*}{Unseen (\%)}  
& SG~\cite{vu2022softgroup} & 0.32 & 0.05 & 0.0 & 3.34 & 0.0 & {11.5} & 1.2 & 6.20 & 1.12 & 2.64 \\ 
                        
& AGP~\cite{liu2022autogpart} & 1.9 & 0.0  & 0.0  &9.3 &0.01  &7.8 &10.02 &6.1  &1.5  &4.07  \\ 

& GAPartNet~\cite{geng2023gapartnet} &2.27 &0.19  &0.0 &10.94  & 0.02  &10.42   &18.32  &14.07  & 3.1 & 6.59 \\ 

& Ours       & \textbf{28.88} & \textbf{0.925} & \textbf{0.67} & \textbf{51.69} & \textbf{1.23} & \textbf{28.53} & {\textbf{20.47}} &\textbf{24.05} &\textbf{18}  &\textbf{19.38}  \\ \bottomrule[1pt]
\end{tabular}
}

\end{table*}

In the pose estimation tasks, the results in Tab.~\ref{tab:pose_evel} demonstrate a significant improvement in pose accuracy with our method compared to the previous state-of-the-art method, GAParNet, particularly in the {\boldmath{$R{e}$}} and $\textbf{A}_{5}$ metric, confirming the accuracy of the estimated poses. GAParNet does not effectively resolve ambiguities in parts like hinged lids or doors; it only uses a relaxed metric that permits 180$^\circ$ mirror symmetry due to its reliance on local part inputs. In our strict evaluation, we remove this tolerance, leading to significant rotation errors in its two-stage method. This two-stage method struggles to distinguish poses based on individual segmented parts, as shown in Fig.~\ref{fig:ambiguity}. This highlights the importance of considering global context.

\begin{figure*}
    \centering
    \includegraphics[width=0.9\textwidth]{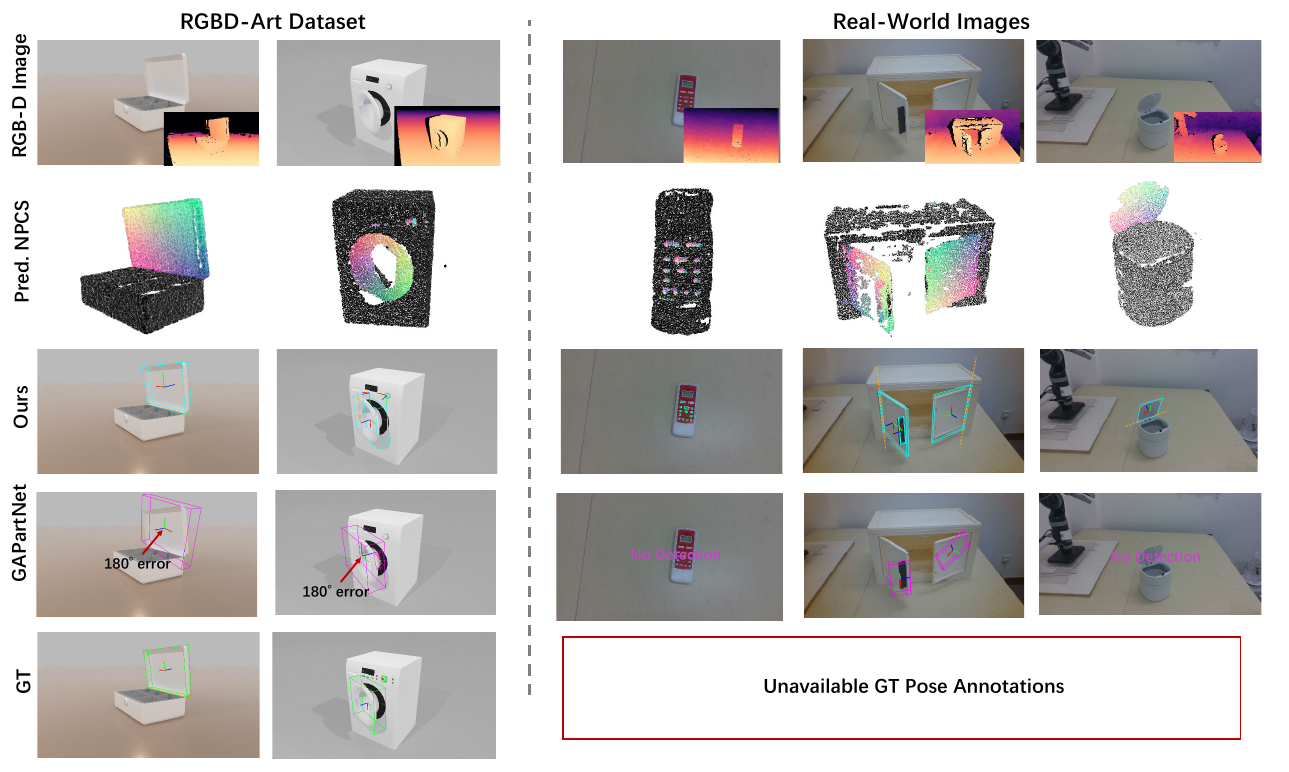}
        \vspace{-0.1in}   
    \caption{Qualitative results from the RGBD-Art dataset and real-world images captured using the RealSense D435 camera. We showcase the RGB-D images, the estimated NPCS for each component of our method, and the pose and size results. Additionally, we provide comparisons with the baseline method GAPartNet~\cite{geng2023gapartnet} and ground truth annotations.\label{fig:compare} }
 
    \vspace{-0.15in}   
\end{figure*}

We also present qualitative results in Fig.~\ref{fig:compare}, which demonstrate the sim-to-real capability of our method, even with noisy depth data from a consumer-grade RealSense camera in real-world settings. In contrast, the point-based method GAPartNet suffers a drop in accuracy. 
Unlike the two-stage method GAParNet, our single-shot approach considers the entire object input rather than just the segmented part. This helps to resolve visual ambiguity in pose identification, as shown in Fig~\ref{fig:ambiguity} and Fig.~\ref{fig:compare}. We also present comparison results for indoor large scene parts from the real-world OPD~\cite{jiang2022opd} and MultiScan~\cite{mao2022multiscan} datasets. For more results, please refer to the Appendix.

\begin{table}[htb] 
\centering
\footnotesize
\setlength\tabcolsep{4pt}
\caption{{Results of part pose estimation on different evaluation metric.} PG=baseline modified from PointGroup~\cite{jiang2020pointgroup}. AGP=baseline modified from AutoGPart~\cite{liu2022autogpart}.\label{tab:pose_evel}}
\vspace{-0.1in}
\begin{tabular}{c|cccccc}
\toprule
Method & {\boldmath{$R_{e}$}} $\downarrow$ & {\boldmath{$T_{e}$}}$\downarrow$ & {\boldmath{$S_{e}$}}$\downarrow$ & \textbf{mIoU} $\uparrow$ & $\textbf{A}_{5}$ $\uparrow$ & $\textbf{A}_{10}$ $\uparrow$\\
\midrule[0.5pt]
PG~\cite{jiang2020pointgroup}  & 89.30 & 0.091 & 0.057 & 18.41 & 0.54 &  1.21\\ 
AGP~\cite{liu2022autogpart}  & 99.40 & 0.099 & 0.061 & 20.10 & 0.53 & 1.32 \\
GAPartNet~\cite{geng2023gapartnet}  & 83.3 & {0.061} & {0.043} & 39.53 & 0.71 & 1.40 \\
Ours  & \textbf{10.39} & \textbf{0.055} & \textbf{0.026}  & \textbf{56.23} & \textbf{33.91} & \textbf{58.44} \\ 
\midrule[0.5pt]
\end{tabular}
\end{table}

\subsection{Ablation study}
\label{sec:ablation}
To evaluate the effectiveness of each component in our framework, we conduct an ablation study to evaluate the contributions of multi-task training and RGB features incorporation. The results of these evaluations are summarized in Tab.~\ref{tab:ablation_cotrain} and Tab.~\ref{tab:ablation_rgb}.

\noindent\textbf{Importance of multi-task learning.}
We trained three individual sub-modules of our method, each supervised by a separate task. After training, we combined the results from the independently trained heads to estimate the final pose and size, and compared this performance with that of the model trained using multi-task learning. As shown in Table~\ref{tab:ablation_cotrain}, the multi-task model consistently outperforms the individual models across all metrics. 
These results show that multi-task learning enhances pose and size estimation accuracy. Furthermore, the end-to-end optimization of the three sub-tasks improves performance, unlike a cascaded approach that may lead to error accumulation.

\begin{table}[h] \small
\centering
\footnotesize
\caption{Comparison of model via multi-task training and individual training under different evaluation metric. \label{tab:ablation_cotrain}}
\vspace{-0.1in}
\begin{tabular}{c|cccccccc}
\toprule
Method & {\boldmath{$R_{e}$}} $\downarrow$ & {\boldmath{$T_{e}$}}$\downarrow$ & {\boldmath{$S_{e}$}}$\downarrow$ & \textbf{mIoU} $\uparrow$ & $\textbf{A}_{5}$ $\uparrow$ & $\textbf{A}_{10}$ $\uparrow$\\
\midrule[0.5pt]
Individual & 15.30 & 0.063 & 0.036 & 51.75 & 27.41 & 50.86 \\
Multi-task & \textbf{10.39} & \textbf{0.055} & \textbf{0.026}  & \textbf{56.23} & \textbf{33.91} & \textbf{58.44} \\  
\midrule[0.5pt]
\end{tabular}
\end{table}

\noindent\textbf{Importance of dense RGB features.}
To assess the contribution of RGB features, we start with a baseline method using the PointNet++ backbone, which relies solely on point cloud data. We then sequentially incorporate vision backbones, such as SAM2~\cite{ravi2024sam}, FeatUp~\cite{fu2024featup}, and DINOv2~\cite{oquab2023dinov2}. The comparison results in Tab.~\ref{tab:ablation_rgb} demonstrate that the SAM2 encoder notably improves accuracy. FeatUp, which provides high-resolution features, outperforms DINOv2 and further enhances performance. 
The full model leverages both feature types for optimal results, with RGB features enhancing the segmentation of small parts that are difficult to achieve with point cloud data alone.

\begin{table}[h] \small
\centering
\footnotesize
\setlength\tabcolsep{4pt}
\caption{Comparison of variants of model and our CAP-Net. `PNet' indicates network modified from PointNet++~\cite{qi2017pointnet++}. \label{tab:ablation_rgb}}
\vspace{-0.1in}
\begin{tabular}{l|cccccc}
\toprule
Method & {\boldmath{$R_{e}$}} $\downarrow$ & {\boldmath{$T_{e}$}}$\downarrow$ & {\boldmath{$S_{e}$}}$\downarrow$ & \textbf{mIoU} $\uparrow$ & $\textbf{A}_{5}$ $\uparrow$ & $\textbf{A}_{10}$ $\uparrow$ \\
\midrule[0.5pt]
PNet & 21.30 & 0.062 & 0.060 & 41.09 & 16.50 & 38.87 \\
PNet + SAM & 12.20 & 0.063 & 0.027 & 48.79 & 23.68 & 47.90 \\
PNet + FeatUp& 20.03 & 0.058 & 0.054 & 43.52 & 21.44 &  42.81 \\
PNet + DINOv2  & 25.12 & 0.062 & 0.060 & 40.10 & 13.68 & 34.36 \\
Full Model & \textbf{10.39} & \textbf{0.055} & \textbf{0.026} & \textbf{56.23} & \textbf{33.91} & \textbf{58.44} \\ 
\midrule[0.5pt]
\end{tabular}
\end{table}

\subsection{Robotic Experiment}
\noindent\textbf{Hardware Settings.} Our algorithm is deployed on a PC workstation equipped with an Intel i9-13900K CPU and an NVIDIA RTX 6000 Ada Generation GPU to provide visual perception of the target objects. For executing the grasping and manipulation tasks, we utilize the Kinova Gen2 6-DoF robotic arm. This robotic arm features three under-actuated fingers, each of which can be individually controlled. The RealSense D435 captures RGB-D images from a tripod positioned opposite the robot workspace and is calibrated to the robotic base frame.

\noindent \textbf{Manipulation Strategy.}
Using the NPCS representation, we obtain the joint or prismatic axis and predefined grasp poses in the NPCS frame. By aligning the NPCS with the real-world point cloud, we transform these actionable axes and grasp poses to the camera frame. With camera calibration to the robot's base frame, this allows easy transformation of grasp poses to the robot frame for motion planning. Detailed motion policies for each category of part  refers to our supplemental material.


\noindent\textbf{Task Description.}
To assess the sim-to-real capability of our method and evaluate its robustness and generalizability, we deployed our algorithm onto a real KINOVA robotic arm. To ensure the representativeness of our experiments, we selected three distinct part classes: drawer, hinge lid, and hinge handle. The corresponding tasks involved pulling the drawer, lifting the lid, and raising the handle.

\noindent\textbf{Evaluation Metric.} 
Depending on the specific experimental task, different metrics are used. For the drawer task, the robot arm successfully completed the task by pulling the drawer out by 0.2 meters. For the hinge handle task, success was defined by rotating the handle 30 degrees around its axis. Similarly, for the hinge lid task, the robot arm successfully completed the task by rotating the lid 50 degrees around its axis.

\noindent\textbf{Results.}
The success rate of manipulating articulated objects in real-world robotic experiments is summarized in Table~\ref{tab:comparison}. The results show that our lightweight model competes effectively with the baseline method, GAParNet. Our single-shot approach accurately generates poses that guide the robot in interacting with objects not seen during the training stages, demonstrating the utility of our method in robotic applications. More results please refers to our supplemental material and video.

\begin{table}
\footnotesize
\setlength\tabcolsep{4pt}
    \centering
    \caption{Success Rate of Robot Manipulation. \label{tab:comparison}}
    \vspace{-0.1in}
    \begin{tabular}{ccccccc}
        \toprule
        & Hinge Handle & Drawer & Hinge Lid & Total \\
        \hline
        GAPartNet~\cite{geng2023gapartnet} & 1/10 & {2/10} & 2/10 & 5/30\\
        \midrule[0.5pt]
        Ours & \textbf{9/10} & \textbf{10/10} & \textbf{9/10} & \textbf{28/30} \\
        \bottomrule
    \end{tabular}
    \vspace{-0.15in}
   
\end{table}


\section{Conclusion}
We introduce CAP-Net, a unified model for estimating the 6D pose and size of articulated parts at the category level. Unlike multi-stage methods, CAP-Net uses a single-stage framework on object-level point clouds for end-to-end part pose and size estimation.
Experiments on our RGBD-Art dataset and real-world data validate its synthetic-to-real capability, and robotic tests demonstrate its effectiveness on unseen articulated objects. However, our method relies on depth for pose fitting, and missing surface points can affect pose estimation accuracy. While off-the-shelf depth-completion methods can address this issue, it is beyond the scope of this paper and will be explored in future work. 


\noindent\textbf{Acknowledgment.} 
This work is supported in part by NSFC Project (62176061), Shanghai Municipal Science and Technology Major Project (No.2021SHZDZX0103), Shanghai Technology Development and Entrepreneurship Platform for Neuromorphic and AI SoC, and the Science and Technology Commission of Shanghai Municipality(No. 24511103100). Yanwei Fu and Xiangyang Xue are the corresponding authors.


\appendix
\section*{Appendix}
This supplemental material is organized as follows: In Section~\ref{sec:real-world results}, we compare results on real-world datasets and analyze inference speed in Section~\ref{sec:infer_speed}. Section~\ref{sec:dataset} includes additional examples from our RGBD-Art dataset. Details of the robotic experiments can be found in Section~\ref{sec:robot}. Section~\ref{sec:unseen} presents and analyzes pose and size estimation results for unseen objects, small parts, and symmetric parts. Finally, we showcase more qualitative results to demonstrate the accuracy and sim-to-real adaptability of our approach in Section~\ref{sec:vis}.


\section{Results on Real-world Dataset} 
\label{sec:real-world results}
OPD~\cite{jiang2022opd} and MultiScan~\cite{mao2022multiscan} datasets are valuable contributions to the study of articulated objects, focusing on large scene-level parts with relatively coarse pose annotations while lacking smaller parts and fine-grained pose and size annotations.
While our method should perform well to these datasets, they are not our main focus due to their limited coverage of smaller parts. We have attempted to use the MultiScan dataset, but its depth images remain inaccessible due to the `.zlib' depth format, which we have been unable to decode. We provide motion axis error results on OPD-real dataset as in Tab.~\ref{tab:opd}, highlighting robustness to realistic distorted depth.

\begin{table}[ht]
\centering
\footnotesize
\renewcommand\arraystretch{0.1}
\setlength{\tabcolsep}{2pt} 
\caption{Comparison results on OPD~\cite{jiang2022opd} dataset.\label{tab:opd}}
\vspace{-0.1in}
\begin{tabular}{c|cccccc}
\toprule[1pt]
Method      & SG  & AGP  & GAPNet & OPD-C & OPD-O & Ours \\
\midrule[0.5pt]
Motion axis error$\downarrow$ & $11.21^\circ$ & $12.03^\circ$ & $6.31^\circ$ & 9.06$^\circ$  & 6.67$^\circ$ & \textbf{5.47$^\circ$}  \\
\bottomrule[1pt]
\vspace{-0.15in}
\end{tabular}
\end{table}

\begin{figure}[h]
    \centering
    \vspace{-0.15in}
    \includegraphics[width=1.0\linewidth]{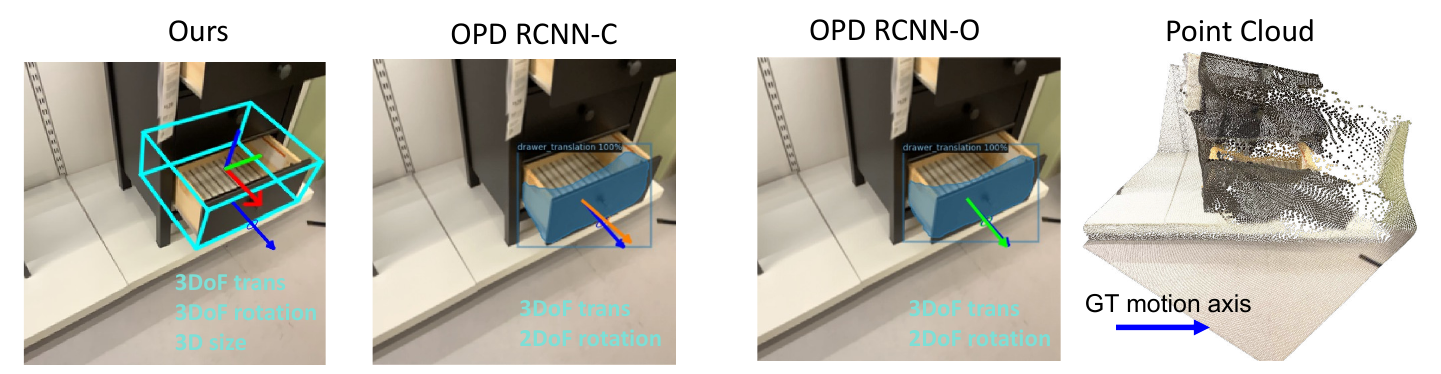}
\vspace{-0.15in}
\end{figure}

\section{Inference Efficiency Analysis}
\label{sec:infer_speed}
The pre-trained vision model (SAM~\cite{ravi2024sam} and FeatUp~\cite{fu2024featup}) slightly increases computational cost compared to the baselines (tested on an A6000 GPU), as shown in the Tab.~\ref{tab:infer_speed}. However, this cost is manageable for realistic robotics tasks, as demonstrated in the video. Inference speed is not the primary issue. During training, the backbone can preprocess images and store feature embeddings to save time. We also plan to optimize inference speed through distillation or quantization methods in future work.

\begin{table}[ht]
\centering
\footnotesize
\renewcommand\arraystretch{0.2}
\caption{Inference speed of different methods.\label{tab:infer_speed}}
\vspace{-0.1in}
\begin{tabular}{c|cccc}
\toprule[1pt]
Method      & SG  & AGP  & GAPartNet & Ours \\
\midrule[0.5pt]
Inference (Hz) & 5 & 7 & \textbf{15} & 4  \\
\bottomrule[1pt]
\vspace{-0.15in}
\end{tabular}
\end{table}

\section{Dataset Examples}
\label{sec:dataset}
\begin{figure*}
    \centering
    \includegraphics[width=0.7\textwidth]{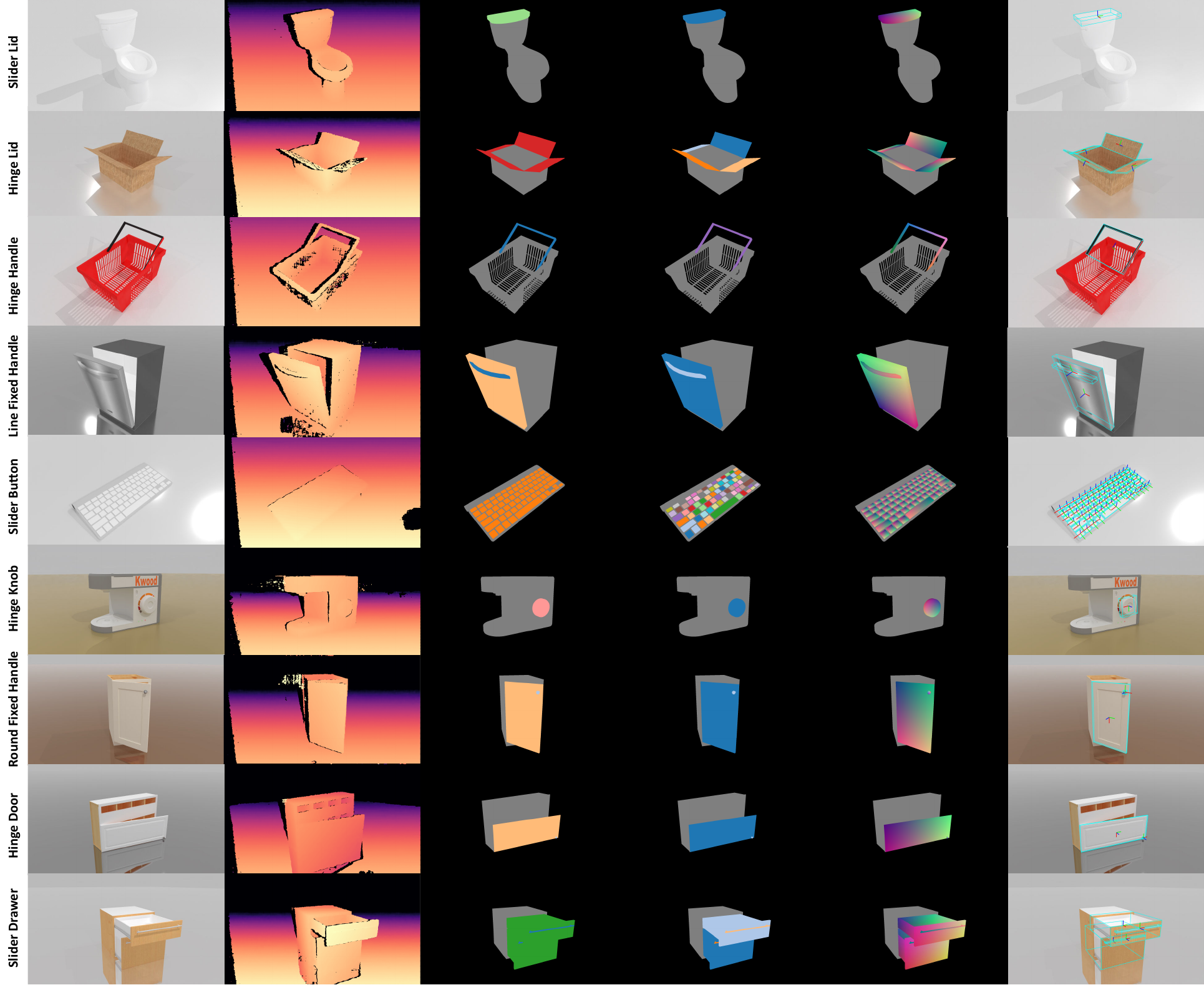}
        \vspace{-0.1in}   
    \caption{\textbf{Seen examples both used for training and testing in our RGBD-Art dataset.} We show the photo-realistic RGB image, realistic depth images, corresponding ground-truth annotations of semantic label, instance label, NPCS map, 6D pose and size.\label{fig:examplar_seen_more} }
\end{figure*}

\begin{figure*}[h]
    \centering
    \includegraphics[width=0.7\textwidth]{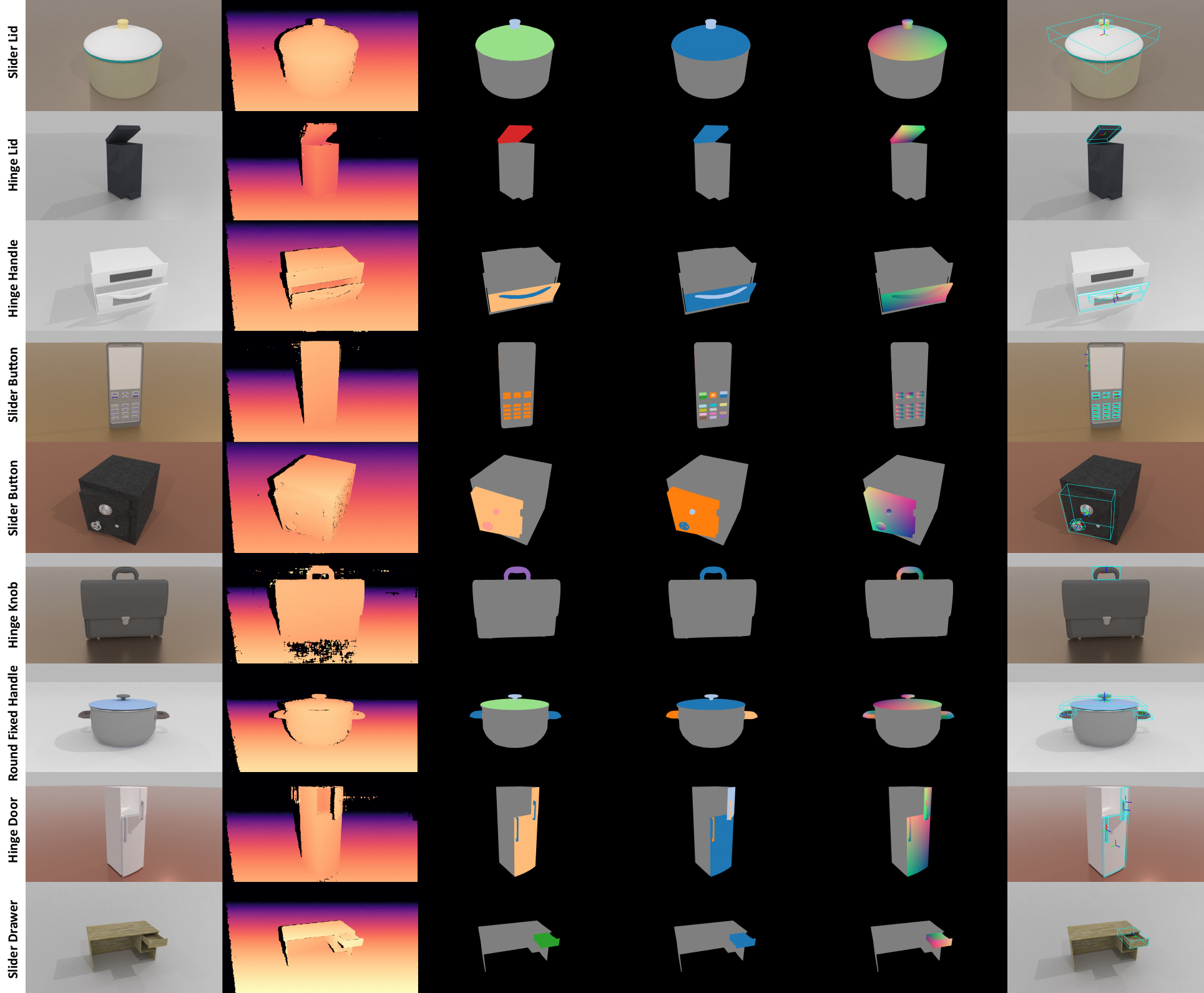}
        \vspace{-0.1in}   
    \caption{\textbf{Unseen examples used for testing in our RGBD-Art dataset.} We show the photo-realistic RGB image, realistic depth images, corresponding ground-truth annotations of semantic label, instance label, NPCS map, 6D pose and size.\label{fig:exemplar_unseen_more} }
    \vspace{-0.15in}   
\end{figure*}

We present additional rendered RGB-D images and their corresponding annotations from our RGBD-Art dataset in Fig.~\ref{fig:examplar_seen_more} and Fig.~\ref{fig:exemplar_unseen_more}. The dataset is divided into two subsets: \textbf{seen} (Fig.~\ref{fig:examplar_seen_more}) and \textbf{unseen} (Fig.~\ref{fig:exemplar_unseen_more}). The \textbf{seen} subset contains objects with articulated parts similar to those in the training categories, while the \textbf{unseen} subset includes novel objects with previously unseen articulated parts that belong to the same categories.

\section{Robotics Experiment Setup}
\label{sec:robot}

\begin{figure}[h]
    \centering
    \vspace{-0.15in}
    \includegraphics[width=1.0\linewidth]{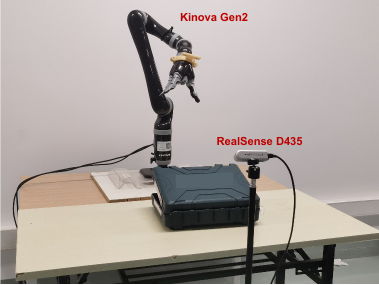}
    \caption{Robotic Setup. \label{fig:robotic_setup}}
\vspace{-0.15in}

\end{figure}

\noindent\textbf{Robotic Setup.}
We use the Kinova Gen2 6-DoF robotic arm to test our algorithm. The RealSense D435 camera captures RGB-D images of the scene and is mounted on a tripod across from the robot's workspace. The camera is calibrated to the robotic base frame, as shown in Figure~\ref{fig:robotic_setup}.

\noindent\textbf{Manipulation Strategy.}
Similar to GAParNet~\cite{geng2023gapartnet}, we adopt the manipulation stratey as follows after estimating the pose and size:
\begin{enumerate}
    \item \textbf{Round Fixed Handle}: Approach the handle along the positive z-axis, open the gripper wider than the bounding box, and then close it to grasp.
    
    \item \textbf{Line Fixed Handle}: Similar to the round handle, but orient the gripper's opening perpendicular to the handle, aligning it with the y-axis of the bounding box.
    
    \item \textbf{Hinge Handle}: Approach and grasp the hinge handle, then rotate it around the predicted axis of the revolute joint.
    
    \item \textbf{Slider Button}: Close the gripper, approach the button from the positive z-axis, and press it.
    
    \item \textbf{Slider Drawer}: Approach an open drawer along the z-axis to retrieve items, or along the x-axis to open it, typically targeting a handle on the front face.
    
    \item \textbf{Hinge Door}: Grab the handle to open the door, rotating the gripper around the predicted shaft. If there’s no handle and the door is ajar, clamp the outer edge along the y-axis to open it.
    
    \item \textbf{Hinge Lid}: Use a similar approach as for the hinge door.
    
\end{enumerate}



\section{More Quantitative Results}
\noindent \textbf{Unseen Object Results.}
\label{sec:unseen}
We present the pose and size results for unseen objects in our RGBD-Art dataset in Table~\ref{tab:pose_unseen}. The results demonstrate that our method can generalize across object categories, effectively handling novel parts that belong to previously seen categories.

\begin{table}[htb] 
\centering
\footnotesize
\setlength\tabcolsep{4pt}
\caption{{Results of part pose estimation on unseen object categories.} PG=baseline modified from PointGroup \cite{jiang2020pointgroup}. AGP=baseline modified from AutoGPart\cite{liu2022autogpart}.\label{tab:pose_unseen}}
\vspace{-0.1in}
\begin{tabular}{c|cccccc}
\toprule
Method & {\boldmath{$R_{e}$}} $\downarrow$ & {\boldmath{$T_{e}$}}$\downarrow$ & {\boldmath{$S_{e}$}}$\downarrow$ & \textbf{mIoU} $\uparrow$ & $\textbf{A}_{5}$ $\uparrow$ & $\textbf{A}_{10}$ $\uparrow$\\
\midrule[0.5pt]
PG~\cite{jiang2020pointgroup}  & 99.78 & 0.131 & 0.091 & 9.63 & 0.34 & 0.56 \\ 
AGP~\cite{liu2022autogpart}  & 105.62 & 0.125 & 0.088 & 12.54 & 0.37 & 0.74 \\
GAPartNet~\cite{geng2023gapartnet}  & 90.81 & 0.073 & 0.052 & 30.71 & 0.54 & 1.03 \\
Ours  & \textbf{12.79} & \textbf{0.062} & \textbf{0.036}  & \textbf{50.54} & \textbf{25.23} & \textbf{50.71} \\ 
\midrule[0.5pt]
\end{tabular}
\end{table}

\noindent \textbf{Improvement on Small-part Objects.}
Table~\ref{tab:small_part} has shown improved performance for small part classes like Hg.Kb and Sd.Bn. We also evaluate performance using an extreme challenge metric of  
$\frac{\text{part \ diameter} }{\text{object  \ diameter}}\leq 0.1 $ to further highlight small part performance. The improved performance on small parts is presented in the table below.

\begin{table}[ht]
\centering
\footnotesize
\setlength{\tabcolsep}{0.5pt} 
\caption{{Comparison results on small parts.} SG=SoftGroup~\cite{vu2022softgroup}. AGP=baseline modified from AutoGPart~\cite{liu2022autogpart}. `-' indicates no detection, and we show only 5 detected classes.\label{tab:small_part}}
\vspace{-0.1in}
\begin{tabular}{c|c|ccccc}
\toprule[1pt]
\textbf{Small Parts(AP50$\uparrow$)} & Method & Ln.F.Hl. & Rd.F.Hl. & Sd.Bn & Hg.Dr. & Hg.Kb.\\
\midrule[0.5pt]
\multirow{4}{*}{Seen}  
& SG~\cite{vu2022softgroup}      & -     & -     & -     & -     & -         \\
& AGP~\cite{liu2022autogpart}     & -     & -     & -     & -     & -         \\
& GAPNet~\cite{geng2023gapartnet}  & -     & -     & 5.92  & -    & -         \\
& Ours    & \textbf{16.36} & \textbf{16.07} & \textbf{38.59} & \textbf{25.00}     & \textbf{0.427}   \\
\midrule[0.5pt]
\multirow{4}{*}{Unseen}  
& SG~\cite{vu2022softgroup}       & -     & -     & -     & -     & -         \\
& AGP~\cite{liu2022autogpart}     & -     & -     & -     & -     & -         \\
& GAPNet~\cite{geng2023gapartnet}  & -     & -     & 8.16  & -     & -         \\
& Ours    & \textbf{13.39} & - & \textbf{21.85}     & \textbf{8.35} &\textbf{0.645}  \\
\bottomrule[1pt]
\vspace{-0.15in}
\end{tabular}
\end{table}

\noindent \textbf{Improvement on Symmetric Parts.} We present per-part pose results for the $R_e$ metric ($\downarrow$), focusing on symmetric parts such as the slider button, hinge door, slider lid, and hinge lid, without symmetry tolerance. The results in Tab~\ref{tab:sym_part} show that our method effectively resolves visual ambiguity in rotation by incorporating global context.
 
\begin{table}[ht]
\centering
\footnotesize
\setlength{\tabcolsep}{3pt} 
\caption{{Comparison results on symmetric parts.} \label{tab:sym_part}}
\vspace{-0.1in}
\begin{tabular}{c|c|cccc}
\toprule[1pt]
Sym.($R_e$$\downarrow$) & Method & Hg.Ld. & Sd.Ld. & Sd.Bn & Hg.Dr. \\
\midrule[0.5pt]
\multirow{2}{*}{Seen}  
& GAPNet  &  24.42 & 147.01  & 59.73 & 75.11  \\
& Ours    &  \textbf{12.37} & \textbf{6.10}  &  \textbf{9.69} & \textbf{6.96}  \\
\midrule[0.5pt]
\multirow{2}{*}{Unseen}  
& GAPNet   & 38.47 &  159.21 &  75.88 & 89.27 \\
& Ours  & \textbf{18.00} & \textbf{28.62}  & \textbf{7.00}   &  \textbf{16.53}    \\
\bottomrule[1pt]
\end{tabular}
\label{tab:sym_res}
\end{table}

\section{More Qualitative Results.}
\label{sec:vis}
\begin{figure*}
    \centering
    \includegraphics[width=0.9\textwidth]{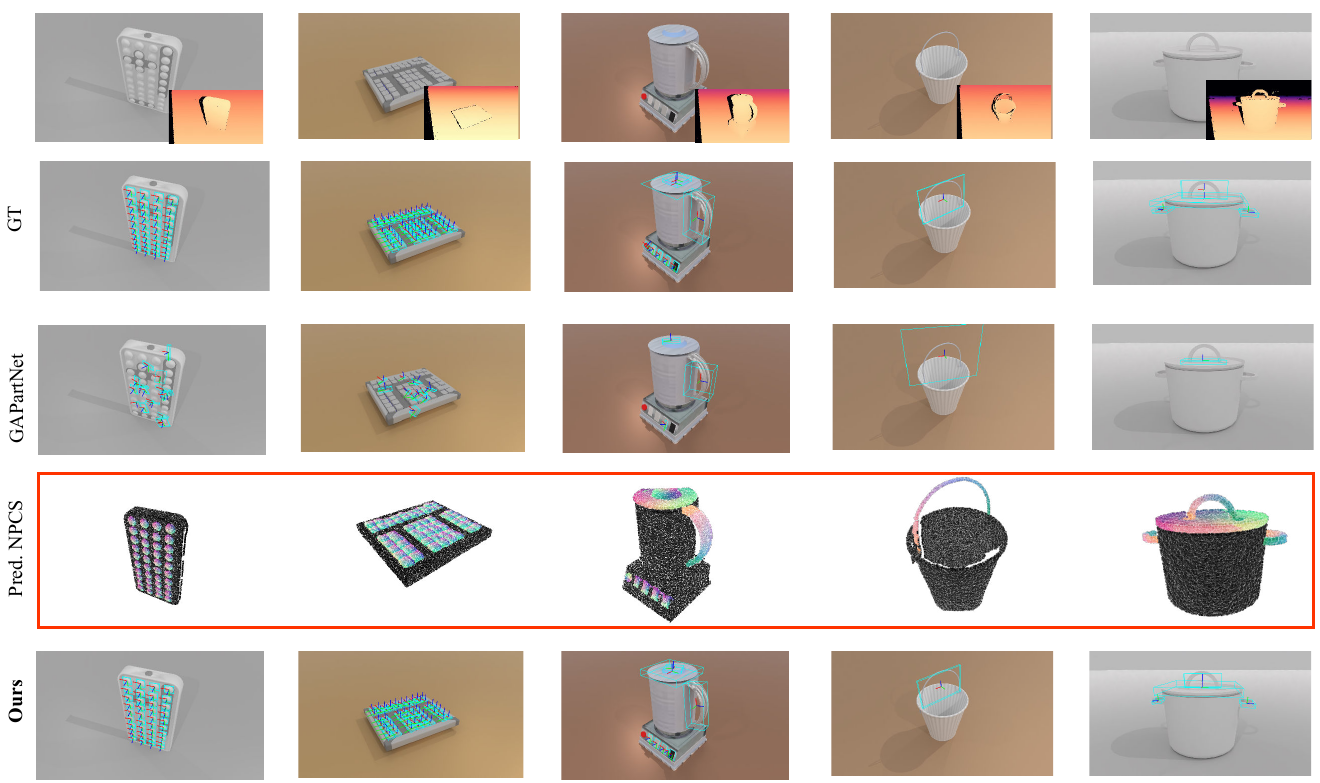}
        \vspace{-0.1in}   
    \caption{\textbf{Qualitative results of the RGBD-Art dataset.} We present the RGB-D images, the estimated NPCS for each component of our method, as well as the resulting pose and size estimations. Additionally, we provide comparisons with the baseline method GAPartNet~\cite{geng2023gapartnet} and ground truth annotations for qualitative evaluation \label{fig:qualitative_dataset}.}
    \vspace{-0.25in}   
\end{figure*}

\begin{figure*}[h]
    \centering
    \includegraphics[width=0.9\textwidth]{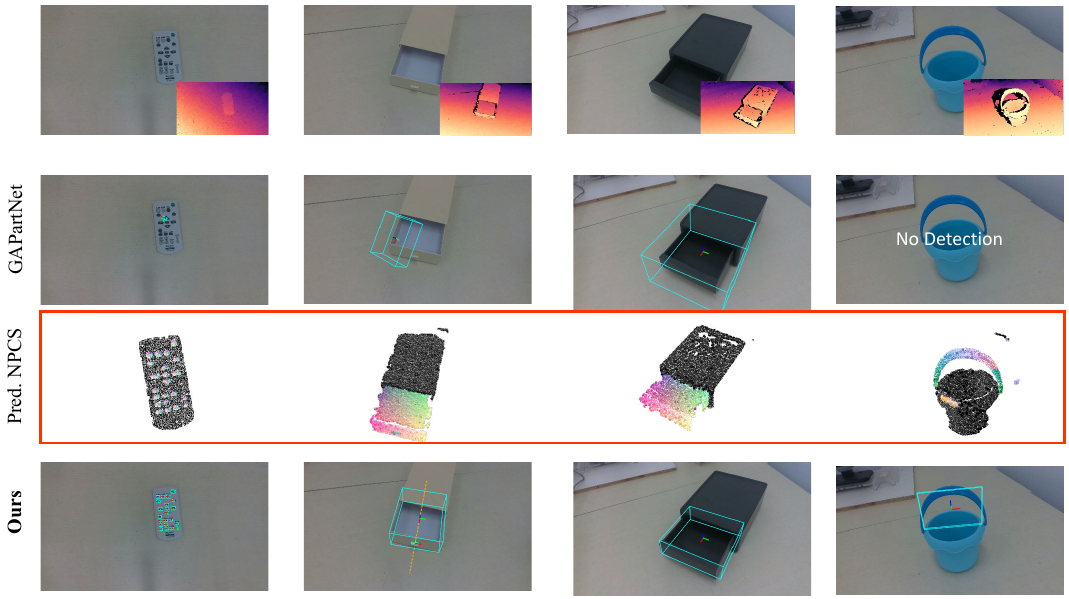}
        \vspace{-0.1in}   
    \caption{\textbf{Qualitative results from real-world images captured using the RealSense D435 camera.} We showcase the RGB-D images, the estimated NPCS for each component of our method, and the resulting pose and size estimations. Additionally, we provide comparisons with the baseline method GAPartNet~\cite{geng2023gapartnet} for qualitative evaluation \label{fig:qualitative_real}.}
\end{figure*}

We present additional qualitative results using our realistic RGBD-Art dataset, which mimics images captured by RealSense D415 and real-world images captured by RealSense D435. These cameras have different baselines, where the baseline of 55mm for the D415 and 50mm for the D435. The results are displayed in Fig.~\ref{fig:qualitative_dataset} and Fig.~\ref{fig:qualitative_real}.

\noindent\textbf{Results on RGBD-Art Datasets.} As shown in Fig.~\ref{fig:qualitative_dataset}, CAP-Net effectively detects small parts and accurately estimates their poses and sizes by leveraging RGB image features.  

\noindent\textbf{Results on Real-world Images.} Despite differences in camera baselines between real-world and training images, the results highlight the sim-to-real capability of our model, demonstrating cross-camera generalization and sim-to-real performance. This also underscores the value of our realism-enhanced dataset. 

\clearpage
\clearpage
{
    \small
    \bibliographystyle{ieeenat_fullname}
    \bibliography{arXiv/egbib}
}


\end{document}